\documentclass[runningheads]{llncs}
\usepackage[utf8]{inputenc}
\usepackage[T1]{fontenc}
\usepackage{graphicx}% Include figure files
\usepackage{dcolumn}% Align table columns on decimal point
\usepackage{bm}% bold math
\usepackage{hyperref}% add hypertext capabilities 
\usepackage{braket}
\usepackage{amsmath}
\hypersetup{
    colorlinks=true,
    linkcolor=blue,
    filecolor=magenta,
    urlcolor=cyan,
    citecolor=blue,
}
\usepackage{cite}
\usepackage{amssymb}
\usepackage{booktabs}
\usepackage{xcolor}
\usepackage{url}
\title{The production of meaning in the processing of natural language}
\titlerunning{The production of meaning in NLP}

\author{
Christopher J. Agostino\inst{1} \and
Quan Le Thien\inst{2} \and
Nayan D'Souza\inst{1,3} \and
Louis van der Elst\inst{4}
}

\authorrunning{C. Agostino et al.}

\institute{
NPC Worldwide, Bloomington, IN, USA \\
\email{cjp.agostino@gmail.com} \and
Department of Physics, Indiana University, Bloomington, IN, USA \and
Department of Linguistics, Indiana University, Bloomington, IN, USA \and
Imperial College London, London, UK
}

\begin{document}
\maketitle

\begin{abstract}

Understanding the fundamental mechanisms governing the production of meaning in the processing of natural language is critical for designing safe, thoughtful, engaging, and empowering human-agent interactions. If meaning is constituted rather than retrieved, then the search for context-independent features or circuits in the pursuit of mechanistic interpretability may be fundamentally limited. Experiments in cognitive science and social psychology have demonstrated that human semantic processing exhibits contextuality more consistent with quantum logical mechanisms than classical Boolean theories, and recent works have found similar results in large language models---in particular, clear violations of the Bell inequality in experiments of contextuality during interpretation of ambiguous expressions. In this work, we explore the CHSH $|S|$ parameter---the metric associated with the inequality---across the inference parameter space of models spanning four orders of magnitude in scale and cross-reference our findings with MMLU, hallucination rate, and nonsense detection benchmarks. We find that the interquartile range of the $|S|$ distribution is completely orthogonal to all external benchmarks, while overall violation rate shows weak anticorrelation with all three benchmarks. We investigate how $|S|$ varies with sampling parameters and word order, and discuss the information-theoretic constraints that genuine contextuality imposes on prompt injection defenses and its human analogue, whereby careful construction and maintenance of social contextuality can be carried out at scale, shaping the space of possible interpretations before any particular one is reached. We consider the implications for mechanistic interpretability and how genuine contextuality sets an information-theoretic bound on the decomposability of semantic processing, such that no context-independent assignment of meanings to internal representations can fully account for interpretive behavior.

\end{abstract}


\section{Introduction}

The fifth-century grammarian Bhart\d{r}hari described the production of meaning as a \textit{spho\d{t}a}, a bursting forth, something grasped whole through \textit{pratibh\={a}}, an intuitive flash that occurs in the listener before any analysis of parts~\cite{bhartrhari}. The schools that opposed him, the M\={\i}m\={a}\d{m}s\={a} and the Ny\={a}ya, did so on the grounds that meaning must be compositionally derived, that the whole is nothing the parts do not already contain~\cite{bhartrhari}. In the 19th century, the Swiss structural linguist Ferdinand de Saussure, who studied Sanskrit writings that themselves were likely influenced by Bhart\d{r}hari's contributions, produced a formulation of the sign as an inseparable unity of signifier and signified, bearing some resemblance to the notion of spho\d{t}a~\cite{saussure}.

Saussure distinguished between two systems: langue and parole. Langue is the abstract formal structure of a language, the grammar and rules that exist independent of any speaker whereas parole is language as it is actually spoken and understood in context. While Saussure treated both as essential, the tradition that followed in the study of language was more aligned with langue than with parole. The assumption that a system can be decomposed into independently analyzable parts whose contributions can be separated and recombined has served as a practical methodological foundation for empirical science since Bacon and Descartes. The nineteenth-century German logician Gottlob Frege adopted this framework for the study of meaning itself when he argued that the meaning of a complex expression is determined by the meanings of its constituent parts and the rules by which they combine~\cite{frege1892,montague1970}. In concert, Zellig Harris proposed that the distributional properties of words in a corpus could serve as a proxy for their semantic relationships~\cite{harris1954}, and these ideas carried through the 20th century and into the 21st, inherited by modern computational linguists in latent semantic analysis and word embeddings~\cite{mikolov2013}. As the utility of distributional semantic models in tasks like information retrieval, translation, and classification became apparent, they were adopted widely across industry, and the deep learning revolution extended their capabilities dramatically. The difficulties that have accompanied the deployment of their latest iteration in large language models---hallucination~\cite{holtzman2020} and susceptibility to prompt injection---persist across models of vastly different scale, architecture, and training. While these difficulties are typically attributed to shortcomings in data, training, or neural network architecture, it is worth considering how the framing of both problems itself stems from the Fregian assumption compositionality in semantic expressions. As commonly understood, hallucinations are assertions made by a language model that are considered verifiably false with respect to some `correct' interpretation; prompt injections are instances where bad actors create a contextual situation wherein the language model is manipulated to maliciously extract sensitive information or to circumvent restrictions. In both cases, the difficulty presupposes that the language model ought to have acted in a specific way that retrieves some pre-established meaning rather than produces it in the act of interpretation. Whether this presupposition is adequate for the study of meaning---a domain in which the observer, the context, and the act of interpretation may not be separable from the thing being studied---has not been formally established~\cite{wittgenstein1953,quine1960,foucault1966,gadamer1960}.

Whether the properties under study are determinate prior to observation or constituted by it is not a question unique to linguistics. The same question arose in twentieth-century physics, where Heisenberg argued that the act of measurement necessarily disturbs the system being observed, Bohr maintained that physical quantities are genuinely indeterminate prior to measurement, and Einstein, Podolsky, and Rosen attempted to demonstrate that quantum mechanics must be incomplete on the grounds that it could not accommodate the simultaneous reality of correlated but spatially separated quantities in entangled particles~\cite{einstein1935}. Bohr replied, but the argument was not widely understood and most physicists were not concerned with what appeared to be a metaphysical question with no bearing on the practical successes of the theory. In 1964, the physicist John Bell derived an inequality that converted the question into an empirical one: any theory in which measured quantities have pre-existing, context-independent values imposes strict limits on the correlations those measurements can produce~\cite{bell1964}. In 1969, John Clauser, Michael Horne, Abner Shimony, and Richard Holt developed a practical test for this, the CHSH test, where $S$ is a sum of correlations between measurement outcomes under different contextual settings~\cite{clauser1969}: $|S| \leq 2$ indicates that the correlations are classical and $|S| > 2$ that they are not. When such a violation occurs, the system's behavior is better described by quantum logic than by classical probability. The application of quantum formalism to cognitive and linguistic systems does not require that these systems operate by quantum mechanical processes; instead, it requires that the structural features the formalism describes---contextuality, non-commutativity, observer-dependence of outcomes---are present regardless of substrate. Bruza et al.~\cite{bruza2023} drew a critical distinction between context-sensitivity, a standard causal influence of context on outcomes, and true contextuality, an acausal form of context-dependence in which a property is genuinely indeterminate prior to the act of measurement\footnote{For a more thorough exploration of the intricacies of indeterminacy versus uncertainty in measurements, see Karen Barad's \textit{Meeting the Universe Halfway}.}. If interpretation is contextual in this stronger sense, then meanings do not pre-exist the interpretive act, and classical probability theory, which presupposes well-defined values prior to observation, is the wrong formalism. The cognitive capacity enabling agents to navigate vast semantic spaces under ambiguity---dubbed relevance realization---is itself non-algorithmic and irreducible to formal computation, operating through context-sensitive attentional mechanisms that dynamically constitute what counts as relevant rather than selecting from a pre-existing set \cite{vervaeke2013,jaeger2023,trouillas2024}. That such a process would produce non-classical correlations when probed experimentally is not surprising; it is precisely the kind of system for which quantum probability theory was developed as a generalization of the classical case.

When cognitive scientists and social psychologists applied Bell's tests in human experiments, they found correlations that violated the classical bound~\cite{busemeyer2012,pothos2022,busemeyer2025,aerts2018a,aerts2018b,bruza2023,wang2014}. More recently, preliminary contextuality experiments have been carried out using large language models, finding a range of $|S|$ values from 1.2 to 2.8~\cite{agostino2025}, and sheaf-theoretic analyses of transformer embeddings have identified extensive quantum-like contextuality in BERT across millions of contextual instances~\cite{lo2024}. Given that experiments on both human cognition and large language models have demonstrated non-local contextuality as a common feature of natural language interpreters, the CHSH $|S|$ parameter offers a way to characterize the interpretive behavior of these systems that does not depend on decomposing the mechanisms by which they produce meaning, and it importantly offers a method for characterizing LLMs in a way that metrics that focus on intelligence or reasoning do not capture.

In this work, to test whether preliminary findings of Bell violations \cite{agostino2025} for a small subset of models generalize to a broader swath of models, we probe the distribution of $|S|$ values across the inference parameter space of large language models spanning four orders of magnitude in parameter count and explore the variations with language model sampling parameters. We also investigate word-order effects as a test of non-commutativity, cross-reference the $|S|$ distributions with existing benchmarks for hallucination and nonsense detection, and consider what these results reveal about the nature of interpretation in these systems, the design of interactions in which humans and language models construct meaning together, and the limits of context-independent mechanistic decomposition.

\section{Methods}
The experiment follows the semantic Bell test protocol of~\cite{agostino2025}, implemented in the quantum semantic toolkit (\texttt{qstk}). We provide a succinct summary of the techniques here. 

Four independent LLM instances each receive an ambiguous sentence under one of four measurement settings---$A$, $A'$, $B$, or $B'$---with no shared state between them. Agents are conditioned only by a short persona prompt (e.g., ``You are a foreign surgeon'' for $A$ and ``You are a bus driver'' for $A^{'}$; ``You are a Sales Executive'' for $B$ and ``You are haunted by past mistakes'' for $B^{'}$,) and interpret both words of an ambiguous pair (e.g., \emph{bank} and \emph{bat}) embedded in a template sentence. A second call classifies each interpretation against the two most common meanings\footnote{These word pairs were specifically selected because they each have two dominant meanings, in some cases interpretations are made by the models that correspond to ones outside of these primary two, but these are relatively rare in our trials ($<1\%
$),}
mapping outcomes to $\pm 1$ vectors. We test five ambiguous word pairs across three sentence templates, flipping word order across trials to probe non-commutativity. For each model--word pair combination, we sweep a $3 \times 3 \times 3$ grid of inference parameters: temperature $\tau \in \{0.2, 1.0, 1.8\}$, nucleus threshold $p \in \{0.7, 0.9, 1.0\}$, and top-$k \in \{10, 50, 100\}$, running ten trials per configuration~\cite{li2025temp,peeperkorn2024,renze2024,meister2023a,meister2023b}.

The models span four orders of magnitude in parameter count and include both dense transformers and sparse mixture-of-experts architectures, running locally via Ollama or through cloud APIs (Gemini, DeepSeek-V3, GPT-4o/4o-mini, Claude Haiku 4.5, Claude Sonnet 4.6). Claude models were tested at default parameters only due to API restrictions. The CHSH $S$-parameter is computed via density matrix formalism: each trial yields a normalized 4-vector of setting products $\mathbf{v}$, from which $\rho = \frac{1}{N}\sum_i |\mathbf{v}_i\rangle\langle\mathbf{v}_i|$ produces expectation values $E(XY) = 4\,\mathrm{Tr}(\rho\,\hat{O}_{XY})$. The parameter $S = E(A,B) - E(A,B') + E(A',B) + E(A',B')$ is bounded by $|S| \le 2$ classically.

\section{Results}

We collected ${\sim}32{,}700$ trials across thirty-two models and five word pairs, with twenty-five models yielding sufficient data for distributional analysis (Table~\ref{tab:distrib}). Every model produced violations of the classical bound: of 1{,}621 valid grid points, 577 (35.6\%) exceeded $|S| > 2$, with values ranging from ${\sim}0.3$ to $4.0$~\cite{agostino2025}. Extremes exceed the Tsirelson bound because observer-side contextuality is not subject to it~\cite{popescu1994}; these occur mainly in smaller models and may reflect overfitting. Every distribution modes near $S = 2.0$, but models differ sharply in spread.

Models differ sharply in IQR: Claude Sonnet 4.6 spans IQR $= 0.55$ while Qwen3 0.6B clusters at IQR $= 0.14$, a sixfold dynamic range that exceeds that of standard deviation or skewness. This spread is not predicted by scale or architecture~\cite{wei2022,schaeffer2023,du2024,michaud2023}---dense and sparse models overlap---and no single sampling parameter drives contextuality consistently; each model produces its own topography across the $(\tau, p, k)$ grid. Word pair choice contributes as much variance as sampling configuration, and word order modulates interpretation (Appendix~\ref{app:commutativity}).

\begin{figure}[htbp]
\centering
\includegraphics[width=\columnwidth]{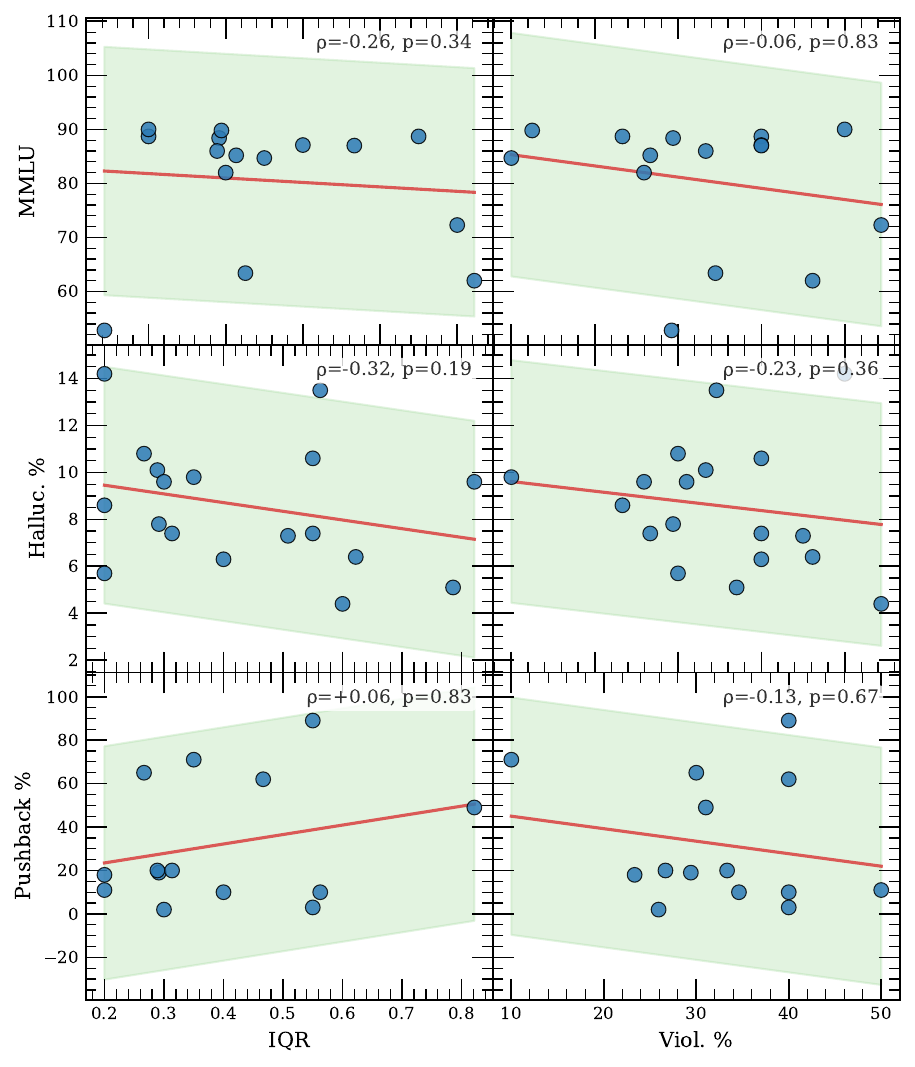}
\caption{$S$ distribution statistics versus external benchmarks. Left: IQR of $S$. Right: violation rate. Rows: MMLU, hallucination rate~\cite{hughes2023}, BullshitBench pushback rate~\cite{bullshitbench2026}. Red lines: linear trends; green bands: $\pm 2\sigma$. The IQR---the most discriminating statistic---shows no relationship to any benchmark.}
\label{fig:benchmarks}
\end{figure}

The IQR and violation fraction of $S$ are orthogonal to existing benchmarks (Figure~\ref{fig:benchmarks}). The IQR correlates negligibly with MMLU ($\rho = -0.10$), hallucination rate ($\rho = -0.27$), and BullshitBench pushback ($\rho = +0.26$); violation rate shows similarly weak anticorrelation ($|\rho| \leq 0.22$).

\begin{figure}[t]
\centering
\includegraphics[width=0.95\columnwidth]{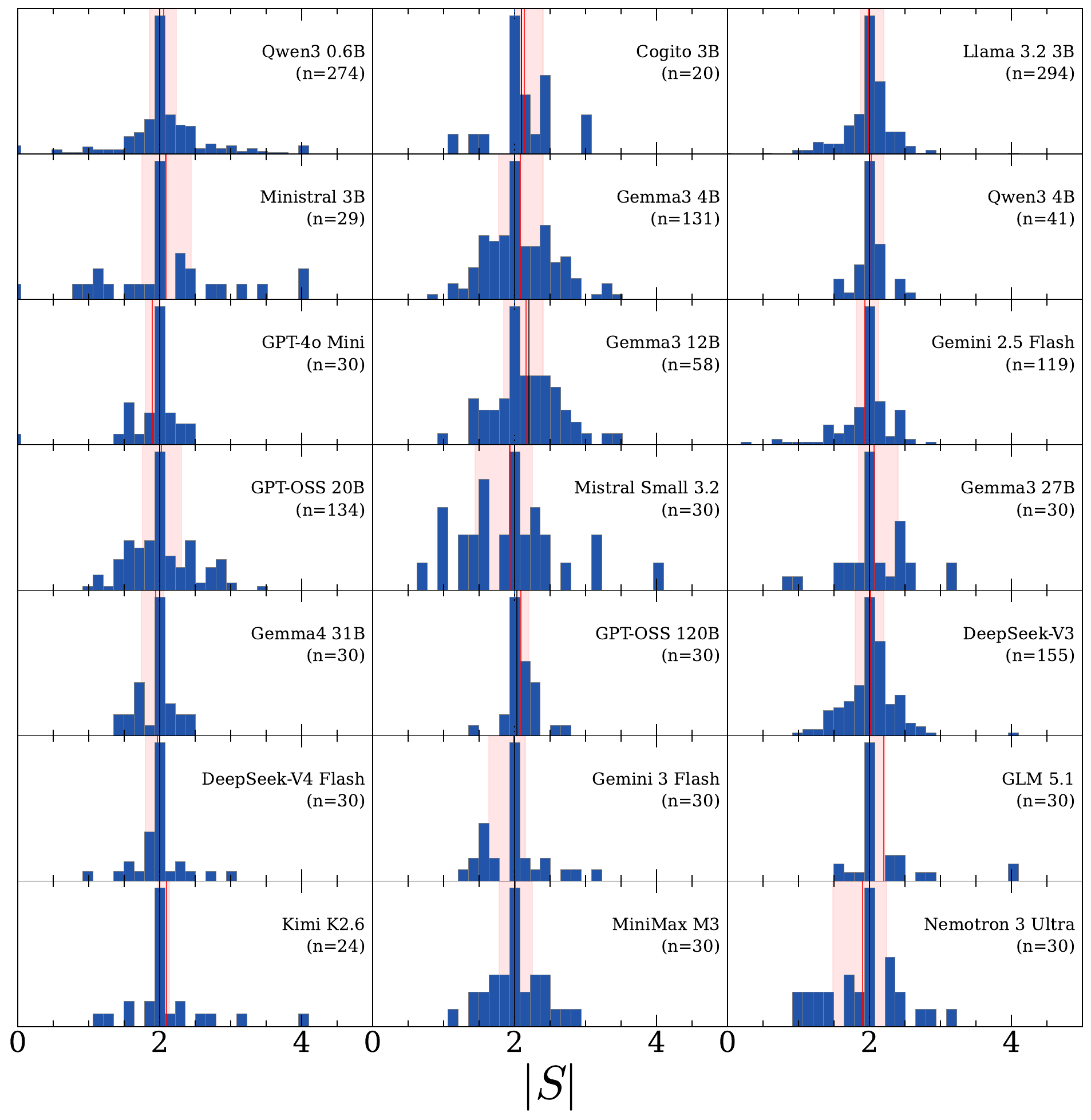}
\caption{Per-grid-point $S$ distributions for each model. Pink bands: IQR. Vertical lines: mean (red), median (black), classical bound $|S|=2$ (dashed blue).}
\label{fig:distrib}
\end{figure}

\section{Discussion}

The distributions we measure show that each model produces a distinct topography of $|S|$ across its parameter space, extending Binz and Schulz's~\cite{binz2023} finding that the character of interpretation is configurable. Two models with similar benchmark scores may behave quite differently as collaborative partners depending on their operating point. The IQR varies by a factor of six across models, and its near-zero correlation with MMLU, hallucination rate, and nonsense-detection benchmarks ($|\rho| \leq 0.27$) suggests that existing evaluation frameworks do not capture the configurability of interpretive character. Models separated by 35 MMLU points can occupy the same IQR range, and models with nearly identical violation rates can differ by a factor of two in IQR. No single scalar skill score predicts how a model distributes its $|S|$ values across conditions.

The Bell violations we observe imply that the frame established by prompt, history, and sampling is constitutive of what is understood, rather than merely conditioning a pre-existing meaning. Mechanistic interpretability methods---activation patching~\cite{wang2022a,makelov2024,heimersheim2024,zhang2026}, dictionary learning~\cite{cunningham2023}, circuit tracing~\cite{conmy2023}---rest on the assumption that internal representations carry context-independent semantic content that can be isolated and inspected. Our results show that this assumption does not hold for the semantic processing measured here, because no assignment of fixed meanings to internal states can reproduce correlations that are shaped by the measurement context in which they arise. The CHSH $|S|$ distribution quantifies the severity of this breakdown and should be used alongside decompositional approaches to mark the conditions under which separability fails. Howard et al.~\cite{howard2014} treat contextuality as a computational resource; the degree of contextuality indicated by a model's $|S|$ distribution may set an upper bound on how much of its interpretive behavior a mechanistic account can capture.

The structural indeterminacy we observe has implications for the design of safety measures. Because the interpretive process is not fully determinate prior to interpretation, a safety filter's pre-classification of intended meaning can be undermined by an adversarial frame that reshapes the measurement context. This asymmetry extends to human cognition, where experiments by Busemeyer et al.~\cite{busemeyer2012}, Bruza et al.~\cite{bruza2023}, and Aerts et al.~\cite{aerts2018a} show that human semantic processing is contextual in the same formal sense. Techniques of propaganda and advertising reshape interpretive context in ways analogous to prompt injection. Where Chomsky and Herman~\cite{chomsky1988} describe the engineering of agreement through information control, our results suggest the engineering of contextuality itself, in which the interpretive frame is shaped before any particular interpretation is reached.

These results also bear on human-AI interaction design. Existing frameworks for assigning automation levels and enabling mixed-initiative interaction~\cite{parasuraman2000,horvitz1999,amershi2019,rezwana2023,ehsan2020,liao2021} do not include a quantitative measure for distinguishing systems that retrieve pre-existing content from systems that co-create meaning through interpretation. If the contextuality identified by Howard et al.~\cite{howard2014} as a resource for quantum computational advantage operates in language models as well, then Bell violation may be a necessary condition for a model to function as a genuine interpretive partner. Because contextuality is configurable and varies with sampling parameters, users and designers would benefit from access to the contextuality profile of their systems, not merely the inference parameters that produce it~\cite{lee2004,roeder2023}.

\section{Conclusions}

In this work, we collected ${\sim}32{,}700$ trials across thirty-two models spanning four orders of magnitude in parameter count. We summarize the main results.

\begin{enumerate}

\item Every model tested produces violations of the classical bound ($|S| > 2$).

\item Each model possesses a distinct contextuality topography across its parameter space. No single parameter drives contextuality consistently, and the relationship to scale is non-monotonic. Architecture and training shape the character of contextuality more than size does.

\item $|S|$ is orthogonal to MMLU, hallucination rate, and nonsense detection benchmarks, capturing a dimension of interpretive behavior that existing evaluations do not address.

\item Word order substantially modulates individual interpretation choices with no directional bias, consistent with non-commuting interpretive observables extending to language models.

\item Genuine contextuality implies that safety cannot be achieved through context-independent guardrails alone, and that the capacity to be manipulated and the capacity to interpret are the same capacity.

\item Semantic interpretation in LLMs is not fully decomposable into context-independent components. The Bell $|S|$ distribution provides a quantitative probe of this non-separability and should be part of the toolkit for characterizing model internals, alongside decompositional methods whose scope and limits it helps to define.

\end{enumerate}
\begin{table}[ht]
\centering
\caption{Distribution statistics of per-grid-point $S$ values for each model (25 models with sufficient data). Each grid point is a unique (word pair, word order, $\tau$, $p$, $k$) configuration. All models mode near the classical bound ($S \approx 2.0$) but differ in spread ($\sigma$), skewness ($\gamma_1$), and excess kurtosis ($\kappa$). IQR = interquartile range. Viol.\% = fraction of grid points with $|S| > 2$. $^\dagger$Tested at default parameters only (no grid sweep) due to API restrictions.}
\label{tab:distrib}
\scriptsize
\begin{tabular}{lccccccc}
\toprule
Model & $n$ & $\sigma$ & $\gamma_1$ & $\kappa$ & IQR & Viol.\% \\
\midrule
Claude Haiku 4.5$^\dagger$   &  10 & 0.43 & $-$1.27 & $+$0.27 & 0.35 & 10.0 \\
Claude Sonnet 4.6$^\dagger$  &  10 & 0.62 & $-$0.97 & $+$1.49 & 0.55 & 40.0 \\
Cogito 3B          &  18 & 0.47 & $-$0.12 & $+$0.47 & 0.40 & 55.6 \\
DeepSeek-V3        & 155 & 0.37 & $+$0.68 & $+$4.99 & 0.40 & 40.0 \\
DeepSeek-V4 Flash  &  30 & 0.36 & $+$0.28 & $+$3.13 & 0.20 & 23.3 \\
Gemini 2.5 Flash   & 119 & 0.37 & $-$1.32 & $+$3.80 & 0.29 & 29.4 \\
Gemini 2.5 Pro     &   8 & 0.40 & $+$1.38 & $+$2.47 & 0.29 & 12.5 \\
Gemini 3 Flash     &  26 & 0.45 & $+$0.65 & $+$0.44 & 0.56 & 34.6 \\
Gemma3 4B          & 130 & 0.49 & $+$0.34 & $+$0.10 & 0.62 & 46.2 \\
Gemma3 12B         &  57 & 0.48 & $+$0.23 & $+$0.50 & 0.60 & 54.4 \\
Gemma3 27B         &  30 & 0.49 & $-$0.27 & $+$1.41 & 0.55 & 40.0 \\
Gemma4 31B         &  30 & 0.26 & $-$0.06 & $-$0.05 & 0.31 & 26.7 \\
GLM 5.1            &  24 & 0.27 & $+$0.86 & $+$1.66 & 0.29 & 33.3 \\
GPT-4o Mini        &  27 & 0.28 & $-$0.09 & $-$0.10 & 0.30 & 25.9 \\
GPT-OSS 20B        & 127 & 0.46 & $+$0.41 & $+$0.15 & 0.58 & 39.4 \\
GPT-OSS 120B       &  30 & 0.24 & $-$0.15 & $+$3.23 & 0.20 & 50.0 \\
Kimi K2.6          &  20 & 0.45 & $+$0.36 & $+$1.15 & 0.27 & 30.0 \\
Kimi K2.7          &  12 & 0.35 & $+$1.73 & $+$4.04 & 0.43 & 41.7 \\
Llama 3.2          & 287 & 0.31 & $-$0.78 & $+$2.27 & 0.33 & 34.5 \\
MiniMax M3         &  30 & 0.39 & $-$0.10 & $+$0.05 & 0.47 & 40.0 \\
Ministral 3B       &  20 & 0.64 & $+$0.02 & $+$0.38 & 0.51 & 45.0 \\
Mistral Small 3.2  &  27 & 0.59 & $+$0.40 & $-$0.15 & 0.79 & 37.0 \\
Nemotron 3 Ultra   &  29 & 0.55 & $+$0.22 & $-$0.12 & 0.82 & 31.0 \\
Qwen3 0.6B         & 325 & 0.38 & $+$0.64 & $+$6.48 & 0.14 & 29.2 \\
Qwen3 4B           &  40 & 0.22 & $+$0.24 & $+$0.62 & 0.20 & 30.0 \\
\bottomrule
\end{tabular}
\end{table}

\bibliographystyle{splncs04}

\bibliography{refs}

@article{popescu1994,
  author  = {Sandu Popescu and Daniel Rohrlich},
  title   = {Quantum nonlocality as an axiom},
  journal = {Foundations of Physics},
  volume  = {24},
  number  = {3},
  pages   = {379--385},
  year    = {1994}
}

@article{aerts2018a,
  author  = {Diederik Aerts and Jonito {Aerts Argu\"{e}lles} and Lester Beltran and Suzette Geriente and Massimiliano {Sassoli de Bianchi} and Sandro Sozzo and Tomas Veloz},
  title   = {Spin and wind directions~{I}: {I}dentifying entanglement in nature and cognition},
  journal = {Foundations of Science},
  volume  = {23},
  pages   = {323--335},
  year    = {2018}
}

@article{aerts2018b,
  author  = {Diederik Aerts and Jonito {Aerts Argu\"{e}lles} and Lester Beltran and Suzette Geriente and Massimiliano {Sassoli de Bianchi} and Sandro Sozzo and Tomas Veloz},
  title   = {Spin and wind directions~{II}: {A} {B}ell state quantum model},
  journal = {Foundations of Science},
  volume  = {23},
  pages   = {337--365},
  year    = {2018}
}

@inproceedings{agostino2025,
  title     = {A quantum semantic framework for natural language processing},
  author    = {Agostino, Christopher J and {Le Thien}, Quan and Apsel, Molly and Pak, Denizhan and Lesyk, Elina and Majumdar, Ashabari},
  booktitle = {International Conference on Quantum Artificial Intelligence and Natural Language Processing},
  pages     = {134--155},
  year      = {2025},
  organization = {Springer}
}

@inproceedings{amershi2019,
  author    = {Saleema Amershi and others},
  title     = {Guidelines for Human-{AI} Interaction},
  booktitle = {Proc. CHI},
  publisher = {ACM},
  pages     = {Paper~3},
  year      = {2019}
}

@article{atmanspacher2012,
  author  = {Harald Atmanspacher and Harald R\"{o}mer},
  title   = {Order effects in sequential measurements of non-commuting psychological observables},
  journal = {Journal of Mathematical Psychology},
  volume  = {56},
  number  = {4},
  pages   = {274--280},
  year    = {2012}
}

@article{einstein1935,
       author = {{Einstein}, A. and {Podolsky}, B. and {Rosen}, N.},
        title = "{Can Quantum-Mechanical Description of Physical Reality Be Considered Complete?}",
      journal = {Physical Review},
         year = 1935,
        month = may,
       volume = {47},
       number = {10},
        pages = {777-780},
          doi = {10.1103/PhysRev.47.777},
       adsurl = {https://ui.adsabs.harvard.edu/abs/1935PhRv...47..777E}
}

@article{bell1964,
  author    = {Bell, J. S.},
  title     = {On the Einstein Podolsky Rosen paradox},
  journal   = {Physics Physique Fizika},
  volume    = {1},
  issue     = {3},
  pages     = {195--200},
  year      = {1964},
  publisher = {American Physical Society},
  doi       = {10.1103/PhysicsPhysiqueFizika.1.195}
}

@book{bhartrhari,
  author = {Bhart\d{r}hari},
  title  = {V\={a}kyapad\={\i}ya},
  note   = {5th century CE. Critical edition and translation by K. A. Subramania Iyer, Deccan College, Poona, 1965--1977},
  year   = {1965}
}

@article{binz2023,
  author  = {Marcel Binz and Eric Schulz},
  title   = {Using cognitive psychology to understand {GPT}-3},
  journal = {Proceedings of the National Academy of Sciences},
  volume  = {120},
  number  = {6},
  pages   = {e2218523120},
  year    = {2023}
}

@article{bock1980,
  author  = {J. Kathryn Bock and David E. Irwin},
  title   = {Syntactic effects of information availability in sentence production},
  journal = {Journal of Verbal Learning and Verbal Behavior},
  volume  = {19},
  number  = {4},
  pages   = {467--484},
  year    = {1980}
}

@article{bruza2023,
  author  = {P.D. Bruza and L. Fell and P. Hoyte and S. Dehdashti and A. Obeid and A. Gibson and C. Moreira},
  title   = {Contextuality and context-sensitivity in probabilistic models of cognition},
  journal = {Cognitive Psychology},
  volume  = {140},
  pages   = {101529},
  year    = {2023},
  doi     = {10.1016/j.cogpsych.2022.101529}
}

@book{busemeyer2012,
  author    = {Jerome R. Busemeyer and Peter D. Bruza},
  title     = {Quantum Models of Cognition and Decision},
  publisher = {Cambridge University Press},
  year      = {2012}
}

@article{busemeyer2025,
  author  = {Jerome R. Busemeyer and others},
  title   = {An overview of the quantum cognition research program},
  journal = {Psychonomic Bulletin \& Review},
  year    = {2025}
}

@article{clauser1969,
  author    = {Clauser, John F. and Horne, Michael A. and Shimony, Abner and Holt, Richard A.},
  title     = {Proposed Experiment to Test Local Hidden-Variable Theories},
  journal   = {Phys. Rev. Lett.},
  volume    = {23},
  issue     = {15},
  pages     = {880--884},
  year      = {1969},
  publisher = {American Physical Society},
  doi       = {10.1103/PhysRevLett.23.880}
}

@inproceedings{du2024,
  author    = {Zhengxiao Du and others},
  title     = {Understanding emergent abilities of language models from the loss perspective},
  booktitle = {Proc. NeurIPS},
  year      = {2024}
}

@incollection{ehsan2020,
  author    = {Upol Ehsan and Mark O. Riedl},
  title     = {Human-centered explainable {AI}: {T}owards a reflective sociotechnical approach},
  booktitle = {HCI International 2020},
  series    = {LNCS},
  volume    = {12424},
  pages     = {449--466},
  year      = {2020}
}

@book{foucault1966,
  author    = {Michel Foucault},
  title     = {Les mots et les choses},
  publisher = {Gallimard},
  address   = {Paris},
  year      = {1966}
}

@article{frege1892,
  author  = {Gottlob Frege},
  title   = {\"{U}ber {S}inn und {B}edeutung},
  journal = {Zeitschrift f\"{u}r Philosophie und philosophische Kritik},
  volume  = {100},
  pages   = {25--50},
  year    = {1892}
}

@book{gadamer1960,
  author    = {Hans-Georg Gadamer},
  title     = {Wahrheit und {M}ethode},
  publisher = {J. C. B. Mohr},
  address   = {T\"{u}bingen},
  year      = {1960}
}

@article{givon1988,
  author  = {Talmy Giv\'{o}n},
  title   = {The pragmatics of word-order: {P}redictability, importance and attention},
  journal = {Studies in Syntactic Typology},
  volume  = {17},
  pages   = {243--284},
  year    = {1988}
}

@article{harris1954,
  author  = {Harris, Z. S.},
  title   = {Distributional Structure},
  journal = {WORD},
  volume  = {10},
  number  = {2-3},
  pages   = {146--162},
  year    = {1954},
  doi     = {10.1080/00437956.1954.11659520}
}

@inproceedings{holtzman2020,
  author    = {Ari Holtzman and Jan Buys and Li Du and Maxwell Forbes and Yejin Choi},
  title     = {The curious case of neural text degeneration},
  booktitle = {Proc. ICLR},
  year      = {2020}
}

@inproceedings{horvitz1999,
  author    = {Eric Horvitz},
  title     = {Principles of mixed-initiative user interfaces},
  booktitle = {Proc. CHI},
  publisher = {ACM},
  pages     = {159--166},
  year      = {1999}
}

@article{howard2014,
  author  = {Mark Howard and Joel Wallman and Victor Veitch and Joseph Emerson},
  title   = {Contextuality supplies the `magic' for quantum computation},
  journal = {Nature},
  volume  = {510},
  pages   = {351--355},
  year    = {2014}
}

@article{jaeger2023,
  author  = {Johannes Jaeger and Anna Riedl and Alex Djedovic and John Vervaeke and Denis Walsh},
  title   = {Naturalizing Relevance Realization: {W}hy agency and cognition are fundamentally not computational},
  journal = {Frontiers in Psychology},
  volume  = {15},
  pages   = {1362658},
  year    = {2024},
  doi     = {10.3389/fpsyg.2024.1362658}
}

@article{lee2004,
  author  = {John D. Lee and Katrina A. See},
  title   = {Trust in automation: {D}esigning for appropriate reliance},
  journal = {Human Factors},
  volume  = {46},
  number  = {1},
  pages   = {50--80},
  year    = {2004}
}

@article{li2025temp,
  author  = {Dongyang Li and others},
  title   = {Exploring the impact of temperature on large language models: {H}ot or cold?},
  journal = {Procedia Computer Science},
  year    = {2025}
}

@article{liao2021,
  author  = {Q. Vera Liao and Kush R. Varshney},
  title   = {Human-centered explainable {AI}},
  journal = {arXiv preprint arXiv:2110.10790},
  year    = {2021}
}

@article{meister2023a,
  author  = {Clara Meister and others},
  title   = {Locally typical sampling},
  journal = {Transactions of the Association for Computational Linguistics},
  volume  = {11},
  pages   = {102--121},
  year    = {2023}
}

@inproceedings{meister2023b,
  author    = {Clara Meister and others},
  title     = {On the efficacy of sampling adapters},
  booktitle = {Proc. ACL},
  pages     = {1437--1455},
  year      = {2023}
}

@inproceedings{michaud2023,
  author    = {Eric J. Michaud and others},
  title     = {The quantization model of neural scaling},
  booktitle = {Proc. NeurIPS},
  year      = {2023}
}

@inproceedings{mikolov2013,
  author    = {Tomas Mikolov and Kai Chen and Greg Corrado and Jeffrey Dean},
  title     = {Efficient estimation of word representations in vector space},
  booktitle = {Proc. ICLR Workshop},
  year      = {2013}
}

@incollection{montague1970,
  author    = {Richard Montague},
  title     = {Universal grammar},
  booktitle = {Theoria},
  volume    = {36},
  pages     = {373--398},
  year      = {1970}
}

@article{parasuraman2000,
  author  = {Raja Parasuraman and Thomas B. Sheridan and Christopher D. Wickens},
  title   = {A model for types and levels of human interaction with automation},
  journal = {IEEE Transactions on Systems, Man, and Cybernetics---Part A},
  volume  = {30},
  number  = {3},
  pages   = {286--297},
  year    = {2000}
}

@inproceedings{peeperkorn2024,
  author    = {Max Peeperkorn and others},
  title     = {Is temperature the creativity parameter of large language models?},
  booktitle = {Proc. ICCC},
  year      = {2024}
}

@article{pothos2022,
  author  = {Pothos, Emmanuel M. and Busemeyer, Jerome R.},
  title   = {Quantum Cognition},
  journal = {Annual Review of Psychology},
  year    = {2022},
  volume  = {73},
  pages   = {749--778},
  doi     = {10.1146/annurev-psych-033020-123501}
}

@book{quine1960,
  author    = {W. V. O. Quine},
  title     = {Word \& Object},
  publisher = {MIT Press},
  year      = {1960}
}

@article{renze2024,
  author  = {Matthew Renze and Erhan Guven},
  title   = {The effect of sampling temperature on problem solving in large language models},
  journal = {arXiv preprint arXiv:2402.05201},
  year    = {2024}
}

@article{rezwana2023,
  author  = {Jeba Rezwana and Mary Lou Maher},
  title   = {Designing creative {AI} partners with {COFI}},
  journal = {ACM Transactions on Computer-Human Interaction},
  volume  = {30},
  number  = {5},
  pages   = {1--28},
  year    = {2023}
}

@article{roeder2023,
  author  = {Lena Roeder and others},
  title   = {A quantum model of trust calibration in human-{AI} interactions},
  journal = {Entropy},
  volume  = {25},
  pages   = {1362},
  year    = {2023}
}

@book{saussure,
  author    = {Ferdinand de Saussure},
  title     = {Cours de linguistique g\'{e}n\'{e}rale},
  publisher = {Payot},
  address   = {Paris},
  year      = {1916}
}

@inproceedings{schaeffer2023,
  author    = {Rylan Schaeffer and Brando Miranda and Sanmi Koyejo},
  title     = {Are emergent abilities of large language models a mirage?},
  booktitle = {Proc. NeurIPS},
  year      = {2023}
}

@book{trouillas2024,
  author    = {Paul Trouillas},
  title     = {A Quantum Theory of Syntax},
  publisher = {Nova Science Publishers},
  year      = {2024}
}

@inproceedings{vervaeke2013,
  author    = {John Vervaeke and Leonardo Ferraro},
  title     = {Relevance, Meaning and the Cognitive Science of Wisdom},
  year      = {2013}
}

@article{wang2014,
  author  = {Zheng Wang and Tyler Solloway and Richard M. Shiffrin and Jerome R. Busemeyer},
  title   = {Context effects produced by question orders reveal quantum nature of human judgments},
  journal = {Proceedings of the National Academy of Sciences},
  volume  = {111},
  number  = {26},
  pages   = {9431--9436},
  year    = {2014}
}

@article{wei2022,
  author  = {Jason Wei and others},
  title   = {Emergent abilities of large language models},
  journal = {Transactions on Machine Learning Research},
  year    = {2022}
}

@book{wittgenstein1953,
  author    = {Ludwig Wittgenstein},
  title     = {Philosophische {U}ntersuchungen},
  publisher = {Basil Blackwell},
  address   = {Oxford},
  year      = {1953}
}

@book{chomsky1988,
  author    = {Noam Chomsky and Edward S. Herman},
  title     = {Manufacturing Consent: The Political Economy of the Mass Media},
  publisher = {Pantheon Books},
  year      = {1988}
}

@misc{hughes2023,
  author    = {Hughes, Simon and others},
  title     = {{HHEM-2.1-Open}: Hallucination Evaluation Model},
  year      = {2023},
  howpublished = {\url{https://github.com/vectara/hallucination-leaderboard}},
  note      = {Vectara Hallucination Leaderboard, accessed 2026-03-04}
}

@misc{bullshitbench2026,
  author    = {{BullshitBench Contributors}},
  title     = {{BullshitBench} v2: Evaluating Nonsense Premise Detection in Language Models},
  year      = {2026},
  howpublished = {\url{https://bullshitbench.com}},
  note      = {Accessed 2026-03-05}
}

@inproceedings{wang2022a,
  author    = {Kevin Wang and Alexandre Variengien and Arthur Conmy and Buck Shlegeris and Jacob Steinhardt},
  title     = {Interpretability in the Wild: a Circuit for Indirect Object Identification in {GPT-2} small},
  booktitle = {International Conference on Learning Representations (ICLR)},
  year      = {2022}
}

@inproceedings{cunningham2023,
  author    = {Hoagy Cunningham and Aidan Ewart and Logan Riggs and Robert Huben and Lee Sharkey},
  title     = {Sparse Autoencoders Find Highly Interpretable Features in Language Models},
  booktitle = {International Conference on Learning Representations (ICLR)},
  year      = {2023}
}

@inproceedings{conmy2023,
  author    = {Arthur Conmy and Augustine N. Mavor-Parker and Aengus Lynch and Stefan Heimersheim and Adri\`{a} Garriga-Alonso},
  title     = {Towards Automated Circuit Discovery for Mechanistic Interpretability},
  booktitle = {Advances in Neural Information Processing Systems (NeurIPS)},
  year      = {2023}
}

@inproceedings{makelov2024,
  author    = {Aleksandar Makelov and Georg Lange and Neel Nanda},
  title     = {Is This the Subspace You Are Looking for? An Interpretability Illusion for Subspace Activation Patching},
  booktitle = {Proc. ICLR},
  year      = {2024}
}

@misc{heimersheim2024,
  author    = {Stefan Heimersheim and Neel Nanda},
  title     = {How to use and interpret activation patching},
  year      = {2024},
  howpublished = {arXiv preprint arXiv:2404.15255},
  note      = {Available at \url{https://arxiv.org/abs/2404.15255}}
}

@misc{zhang2026,
  author    = {Luyang Zhang and Jialu Wang},
  title     = {When Attribution Patching Lies: Diagnosis and a Second-Order Correction},
  year      = {2026},
  howpublished = {arXiv preprint arXiv:2606.09899},
  note      = {Available at \url{https://arxiv.org/abs/2606.09899}}
}

@misc{lo2024,
  author    = {Kin Ian Lo and Mehrnoosh Sadrzadeh and Shane Mansfield},
  title     = {Quantum-Like Contextuality in Large Language Models},
  year      = {2024},
  howpublished = {arXiv preprint arXiv:2412.16806},
  note      = {Available at \url{https://arxiv.org/abs/2412.16806}}
}

\clearpage

\appendix{Appendix: Word-order effects}
\label{app:commutativity}

\begin{figure}[ht]
\centering
\includegraphics[width=0.95\columnwidth]{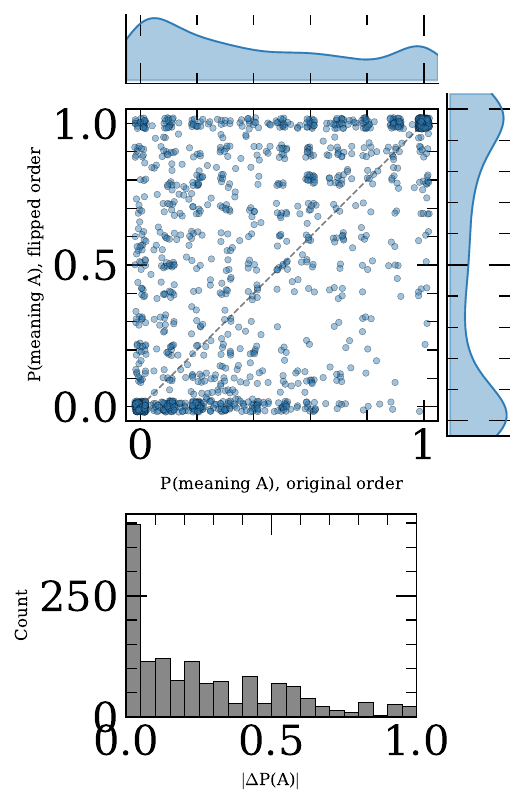}
\caption{Word-order effects on individual interpretation choices. Each point plots $P(\text{meaning A})$ for one word under one interpretive lens and parameter configuration, comparing original versus flipped word order. Points on the diagonal indicate order-invariant interpretation; off-diagonal points indicate non-commutative dependence on presentation order. Marginal densities (top, right) show the bimodal character of interpretation probabilities. Bottom: distribution of $|\Delta P(A)|$ across all paired conditions.}
\label{fig:commutativity}
\end{figure}

Flipping the order of the ambiguous words in the sentence template produces measurable shifts in meaning assignment, consistent with non-commuting interpretive observables~\cite{atmanspacher2012,givon1988,bock1980}. Figure~\ref{fig:commutativity} shows the paired comparison: each point plots $P(\text{meaning A})$ for one word under one interpretive lens and parameter configuration, comparing original versus flipped order. Points on the diagonal indicate order-invariant interpretation; off-diagonal points indicate that presentation order modulates which meaning is selected.

The effect is bidirectional and varies by model and word pair. Many conditions produce $|\Delta P(A)| > 0.5$, indicating that order can flip the dominant interpretation. The marginal densities (top and right panels) show a bimodal structure: outcomes tend to be categorical rather than graded, with models settling decisively on one meaning or the other rather than distributing probability evenly.

Table~\ref{tab:commutativity} gives the per-model, per-word-pair breakdown of the signed difference $S(\text{original}) - S(\text{flipped})$. Positive values indicate higher contextuality in the original order; negative values indicate higher contextuality when flipped. The magnitude quantifies how strongly word order modulates the CHSH parameter for that condition.

\begin{table}[ht]
\centering
\caption{Per-model, per-word-pair commutativity analysis. $\Delta S = S_{\text{orig}} - S_{\text{flip}}$ measures the effect of reversing word order on the CHSH parameter.}
\label{tab:commutativity}
\scriptsize
\begin{tabular}{llccc}
\toprule
Model & Word pair & $S_{\text{orig}}$ & $S_{\text{flip}}$ & $\Delta S$ \\
\midrule
\multicolumn{5}{l}{\textit{(Populated from \texttt{figures/agg\_commutativity.csv})}} \\
\bottomrule
\end{tabular}
\end{table}

\end{document}